\documentclass{article}
\usepackage{spconf,amsmath,graphicx}
\usepackage{subcaption}
\usepackage{pgfplots}
\usepackage{subcaption}
\usepackage{tikz}
\usepackage{pgfplots}
\usepackage{adjustbox}
\usepackage{sidecap}
\pgfplotsset{compat=1.13}
\newcommand\blfootnote[1]{%
  \begingroup
  \renewcommand\thefootnote{}\footnote{#1}%
  \addtocounter{footnote}{-1}%
  \endgroup
}

\title{Conversational Text-to-SQL: An Odyssey into State-of-the-Art and Challenges Ahead}
\name{Sree Hari Krishnan Parthasarathi, Lu Zeng, Dilek Hakkani-T\"ur}
\address{Alexa AI, Amazon}

\begin{document}
\ninept
\maketitle
\begin{abstract}

Conversational, multi-turn, text-to-SQL (CoSQL) tasks map natural language utterances in a dialogue to SQL queries. State-of-the-art (SOTA) systems use large, pre-trained and finetuned language models, such as the T5-family, in conjunction with constrained decoding. With multi-tasking (MT) over coherent tasks with discrete prompts during training, we improve over specialized text-to-SQL T5-family models. Based on Oracle analyses over n-best hypotheses, we apply a query plan model and a schema linking algorithm as rerankers. Combining MT and reranking, our results using T5-3B show absolute accuracy improvements of 1.0\% in exact match and 3.4\% in execution match over a SOTA baseline on CoSQL. While these gains consistently manifest at turn level, context dependent turns are considerably harder. We conduct studies to tease apart errors attributable to domain and compositional generalization, with the latter remaining a challenge for multi-turn conversations, especially in generating SQL with unseen parse trees.\blfootnote{© 2023 IEEE. Personal use of this material is permitted. Permission from IEEE must be obtained for all other uses, in any current or future media, including reprinting/republishing this material for advertising or promotional purposes, creating new collective works, for resale or redistribution to servers or lists, or reuse of any copyrighted component of this work in other works.}

\end{abstract}


\section{Introduction}
\label{sec:intro}

Text-to-SQL is an important research topic in semantic parsing~\cite{dahl1994expanding, zelle1996learning, yu2018spider, yu2019sparc, yu2019cosql, zhong2017seq2sql, ouyang2022training}. Spider~\cite{yu2018spider} and CoSQL~\cite{yu2019cosql} datasets allow for making progress in complex, cross-domain, single and multi-turn text-to-SQL tasks respectively, utilizing a common set of databases, with competitive leaderboards, demonstrating the difficulty in the tasks. In contrast to Spider, CoSQL was collected as entire dialogues, and hence includes additional challenges for the text-to-SQL task in terms of integrating dialogue context. In addition to the challenges in general-purpose code generation~\cite{chen2021evaluating, li2022competition}, where the output of the system is constrained to follow a grammar, the text-to-SQL problem is underspecified without a schema. Since public text-to-SQL tasks use relatively small datasets, previous solutions employ: a) small encoder/decoder models, with constraints on the decoder~\cite{wang2019rat,yin2017syntactic}; b) large pretrained language models (LMs) without constraints on the decoder~\cite{suhr2020exploring, lin2020bridging}, pruning finalized hypotheses. PICARD~\cite{scholak2021picard}, a top entry on the text-to-SQL leaderboard, finetunes a pretrained T5 model and imposes SQL syntax during beam search.

\begin{figure}[ht]
  \centering
  \includegraphics[width=1\linewidth]{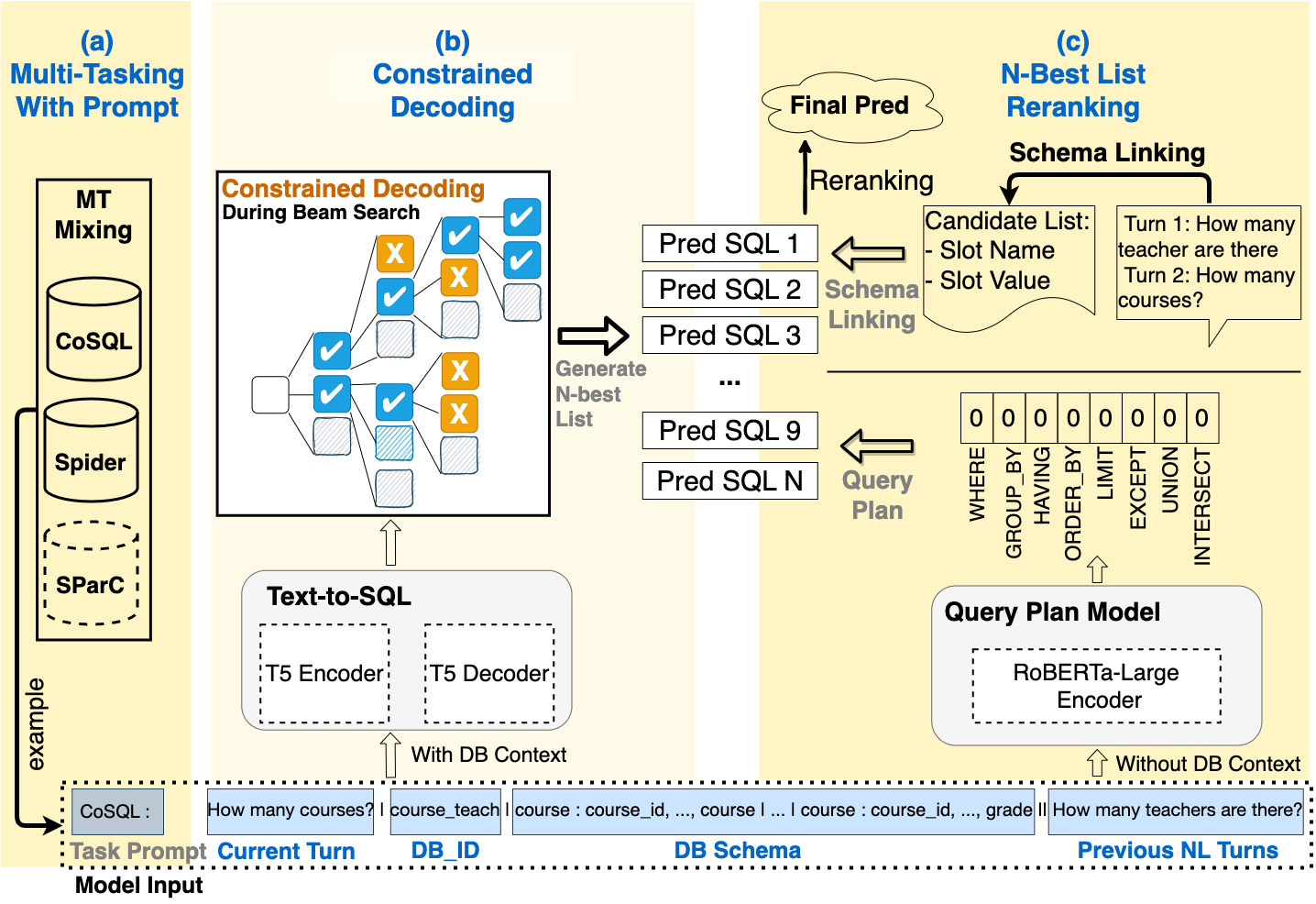}
  \vspace{-6mm}
  \caption{Proposed text-to-SQL system consists of 3 parts: (a) multi-tasking on coherent tasks with discrete prompts; (b) constrained decoding with PICARD; (c) N-best list reranking with SL and QP.}
   \vspace{-2mm}
  \label{fig:proposed_sys}
  \vspace{-3mm}
\end{figure}

Our focus is on multi-turn, conversational text-to-SQL (CoSQL), and we build a system utilizing PICARD. Motivated by the notion that multi-task training (MT)~\cite{bordes2012joint} using coherent tasks and utilizing inductive biases can improve accuracy, we aggregate coherent task data (from CoSQL, Spider, and SParC\footnote{SParC is a multi-turn text-to-SQL dataset, while CoSQL is conversational (both based on the same databases as Spider) -- meaning that there are non-SQL interactions where the system can ask clarifications.}) along with task-specific discrete prompts during training. Next, persuaded by our previous research~\cite{nl2sql_paper}, we conduct Oracle studies on n-best lists from PICARD CoSQL system, inferring that reranking can be helpful. We adapt the two reranking methods from~\cite{nl2sql_paper}, query plan (QP) and schema linking (SL), and show that both methods can help improve multi-turn text-to-SQL. With accuracy on CoSQL being reported using exact-set-match accuracy (EM) and execution accuracy (EX), with T5-Large we observed: a) MT leads to 2.4\% and 1.7\% absolute improvement on EM and EX; b) combined reranking approaches yield 1.9\% and 2.2\% improvements; c) combining MT with reranking, with T5-Large we obtain improvements of 2.1\% in EM and 3.7\% in EX over a T5-Large PICARD baseline. This improvement is consistent on larger models, using T5-3B yielded about 1.0\% in EM and 3.4\% in EX over SOTA baseline. We also submitted our system to the CoSQL leaderboard, our system consistently improves over PICARD baseline on the held out test set on question match (1.2\% absolute) and interaction match (1.1\% absolute). Lastly, we analyze errors in terms of zero-shot domain and compositional generalization. All improvements presented in this paper are absolute gains (i.e., not relative).

\noindent \textbf{Contributions.} The contributions of this paper are: a) proposing MT and combining with n-best reranking to improve over SOTA on a competitive multi-turn, conversational text-to-SQL task; b) analysis of gains at turn level, zero-shot domain generalization, and compositional generalization -- showing challenges in compositional generalization for multi-turn conversations.

\vspace{-2.5mm}
\section{Conversational Text-to-SQL}
\label{sec:nl2sql}

\vspace{-1.5mm}
\subsection{Related Work}
\vspace{-1mm}

\noindent{\textbf{Conversational text-to-SQL}}: A comprehensive survey of text-to-SQL is provided in~\cite{qin2022survey}. Historically, much more research has been undertaken in single-turn text-to-SQL~\cite{dahl1994expanding,yu2018spider,hwang2019comprehensive}. SParC and CoSQL are recent multi-turn text-to-SQL datasets, with CoSQL being more realistic dialogues. Most previous works on multi-turn text-to-SQL attempt to encode context in utterances and/or previous SQL queries into the model~\cite{zhang2019editing,wang2021tracking,cai2020igsql,hui2021dynamic,zheng2022hie}. On the other hand, PICARD~\cite{scholak2021picard} does not treat utterance/SQL context in a special fashion, and relies on large pretrained LMs (PLM) with constrained decoding, obtaining SOTA results. Our work brings context modeling into~\cite{scholak2021picard} in two ways: a) using contextual information in the two n-best reranking methods\cite{nl2sql_paper,hui2021dynamic}; b) better context learning via data augmentation and prompting to reduce cross-task interactions in MT~\cite{aghajanyan2021muppet,xie2022unifiedskg}.

\noindent{\textbf{Compositional generalization}} is an active area of research in semantic parsing~\cite{furrer2020compositional}, with the focus primarily on single-turn utterances~\cite{gu2021beyond}. On the other hand, multi-turn text-to-SQL benchmarks are set up to evaluate cross-domain generalization~\cite{yu2019sparc,yu2019cosql}. In this work, we attempt to bridge both aspects: compositionality analysis in multi-turn cross-domain text-to-SQL.

\vspace{-2mm}
\subsection{Proposed Approach}
\vspace{-1mm} 
The overview of the proposed system is shown in Fig~\ref{fig:proposed_sys}. PICARD is part (b) in the figure, while multi-task prompts and n-best list reranking methods are shown in parts (a) and (c).

\vspace{1mm}

\noindent \textbf{A) Proposed Multi-task Prompt (MT) Approach}: We ensemble data from similar semantic parsing tasks (CoSQL, Spider, and SParC), and inject an inductive bias to the model by using discrete, task-specific, natural language prompts in input. We simply use the task names as the prompts, and the CoSQL example shown in Fig~\ref{fig:proposed_sys} has a prompt  of ``cosql :''.

\vspace{1mm}

\noindent \textbf{B) Proposed Reranking with MT Query Plan (QP) Model}: Starting from~\cite{nl2sql_paper}, we build a model that focuses specifically on improving long span coherence: a multi-label classification model that generates a query plan predicting whether the predicted SQL query should contain any of the 8 clauses (WHERE, EXCEPT, UNION, and INTERSECT, GROUP BY, HAVING, ORDER BY, LIMIT). We adapt the approach so that the QP model is trained in an MT fashion. 

\vspace{1mm}

\noindent \textbf{C) Proposed Reranking with Schema Linking (SL) on conversational context}: We adapt the heuristic algorithm in~\cite{nl2sql_paper} for the multi-turn setting. For each predicted SQL query in n-best list, we follow three steps: a) extract slot names and their respective values from the conditions in the WHERE clause; then check if the slot value exists in any of the referenced tables in the FROM clause. b) obtain a list of candidate slot names/values from the current and previous turns in the interaction, which are exact/partial occurrences of the column/table names and string values in the question with name-based and value-based linking described in RAT-SQL~\cite{wang2019rat}; c) For value linking, we next consider prefix/abbreviation matches on slot values with categorical types.

\vspace{-2mm}
\section{Experimental Setup}
\label{sec:expts}
\vspace{-1.5mm}

\subsection{Datasets and Metrics}
\vspace{-1mm}
We briefly describe the datasets and metrics used in this paper~\cite{yu2018spider}.


\noindent \textbf{A) Text-to-SQL Datasets}: All 3 datasets (CoSQL, Spider, and SParC) are based on 200 databases (covering 138 domains), each with multiple tables. The standard protocol splits them into 140 databases for training, 20 databases for development (DEV), and 40 databases are held back for evaluation. Databases have no overlaps across the splits. CoSQL is a conversational dataset, containing 3k dialogues, with 30k+ turns and 10k+ annotated SQL queries (collected in Wizard-of-Oz fashion). Dialogues are split into 2,164 for training, and 292 for DEV, and 551 for evaluation. Spider is a single-turn dataset, containing 10,181 questions with 5,693 SQL queries. The examples are split into 7,000 for training, 1,034 for DEV, and 2,147 for evaluation. SParC is a sequential text-to-SQL dataset containing 4,298 coherent question sequences; these are split into train (3034), DEV (422), and evaluation (842). We mainly present results on CoSQL (with some discussion on Spider to analyze generalization properties).

\vspace{1mm}

\noindent \textbf{B) Text-to-SQL Metrics}: As mentioned previously, performance is evaluated using EM and EX on the CoSQL and Spider (DEV), with DEV being used for eval, and no hyperparameter tuning being done on them. EM compares each clause between a prediction and its corresponding groundtruth SQL query. The predicted SQL query is correct only if all of the components match. This metric does not take values into account. EX compares the execution output of the predicted SQL query and its corresponding groundtruth SQL queries. Note that both EM and EX can lead to false positives and false negatives. 

\vspace{-2mm}
\subsection{Models}
\vspace{-1mm}
\noindent \textbf{A) Baseline Model}: PICARD~\cite{scholak2021picard} is our baseline, and it is trained on Spider and CoSQL; we focus on two T5 model sizes, T5-Large and T5-3B. Input to the model includes current natural language turn, database name, and serialized database schema (\texttt{table\_name : col1 , ... , coln}) with database content, and previous turns from the dialogue in reverse chronological order. During inference, constrained decoding (CD) is integrated into beam search, with a beam size of 10. 

\vspace{0.5mm}

\noindent \textbf{B) Multi-tasking Prompting (MT) T5 Model}: We introduced two changes: a) augment CoSQL and Spider with SParC and weight them equally during training; b) to reduce variance in estimated parameters, we inject inductive biases with task specific discrete prompts, extending the input. We finetune the model on p3dn.24xlarge instances (8 NVIDIA Tesla V100 GPUs) with teacher forcing and cross-entropy loss for 3000 epochs using a batch size of 2000 and a learning rate of $1e^{-4}$.

\vspace{0.5mm}

\noindent \textbf{C) MT Query Plan (QP) Model}: We finetune RoBERTa-Large models with a sequence classification head on p3.2xlarge instances (1 NVIDIA Tesla V100 GPU). We reused the input from MT T5 model (without database content), and output 1-hot encoded to predict labels extracted from groundtruth queries based on the existence of the 8 clauses. Models are finetuned with binary cross entropy loss for 100 epochs using a batch size of 5 and a learning rate of $1e^{-5}$. 

\vspace{-3mm}
\section{Results}
\label{sec:result}
\vspace{-2.3mm}
\subsection{Multi-Tasking Approach}
\vspace{-1mm}
While we follow a restrictive MT strategy (3 coherent tasks, Spider, CoSQL, and SParC), UnifiedSKG~\cite{xie2022unifiedskg} follows a generalist MT strategy of training on 21 structured knowledge grounding (SKG) tasks (including the 3 tasks above). Table~\ref{table:mt_ph_vs_skg} shows results on CoSQL without CD against baseline. Note that the baseline (52.5\%) performs better than UnifiedSKG (51.6\%). Furthermore, the proposed MT approach performs significantly better than both models.  

\begin{table}[h!]
\vspace{-2mm}
\caption{T5-large: MT performance on CoSQL without CD.}
\vspace{-2mm}
\centering
\label{table:mt_ph_vs_skg}
\begin{tabular}{|c|c|}
\hline
MT Methods &  EM\%\\
\hline
Baseline & 52.5 \\
UnifiedSKG MT-P~\cite{xie2022unifiedskg}      &   51.6   \\
Proposed MT        &   54.6   \\
\hline
\end{tabular}
\vspace{-4mm}
\end{table}
\vspace{-2.25mm}

\subsection{Oracle Analysis and Reranking Approaches}
\vspace{-1mm}
We enable CD and perform Oracle analysis on 10-best hypotheses. The study is done on the baseline and proposed MT model at two selected T5 model sizes: T5-Large and T5-3B. In Table~\ref{table:oracle_over_beam_size}, for each row block, the first row is the 1-best, while the other row shows Oracle accuracies. As can be observed, both EM and EX improve significantly: for example, the T5-Large proposed MT model gains 12.6\% and 10.8\% absolute for EM and EX respectively. Similar gains can be seen for the baseline at the same model size, as well as for the T5-3B models, following trends observed in our paper~\cite{nl2sql_paper}. Note that with T5-large models, we observed that with CD the first and third rows in Table 1 yielded an EM of 54.4\% and 56.8\% respectively.

Table~\ref{table:reranking_effect} lists reranking results on 10-best hypotheses obtained from baseline using a T5-Large model. SL contributes 0.4\% and 2.0\% absolute improvement on EM and EX, while QP yields smaller gains, with the gains from SL and QP being additive.

\vspace{-2mm}
\begin{table}[h!]
\caption{Oracle accuracies with beam size of 10. }
\vspace{1mm}
\centering
\label{table:oracle_over_beam_size}
\vspace{-2mm}
\begin{tabular}{|c|cc|cc|l|}
\hline
{Model} & \multicolumn{2}{c|}{T5-Large} &\multicolumn{2}{c|}{T5-3B}& Notes  \\
{}   & EM\%   & EX\% & EM\%   & EX\% & \\
\hline
Baseline   &  54.4  &    63.7   & 57.1      & 66.6 & 1-best \\
       &  67.0  &    75.7   & 68.5      & 76.7 & Oracle\\
\hline
Proposed MT       &   56.8 &    65.4  & 58.3      &   68.7 & 1-best   \\
       &   69.4 &    76.2  & 72.0      &   78.6& Oracle\\
\hline
\end{tabular}
\vspace{-0mm}
\end{table}

\vspace{-3mm}
\begin{table}[h!]
\vspace{-2mm}
\caption{T5-Large: Reranking approaches QP and SL on CoSQL.}
\vspace{-1mm}
\centering
\label{table:reranking_effect}
\begin{tabular}{|l|cc|}
\hline
Methods & EM\%   & EX\%\\
\hline
Baseline           &   54.4  & 63.7 \\
$\quad +$ SL          &   54.8  & 65.7 \\
$\quad +$ QP           &   54.9  & 63.9 \\
$\quad +$ QP $+$ SL     &   55.3  & 65.9 \\
\hline
\end{tabular}
\vspace{-3mm}
\end{table}

\vspace{-2mm}

\vspace{-2mm}
\subsection{Combined Results with MT and Reranking}
\label{sec:combined_results}
\vspace{-1mm}
We present results combining MT and reranking approaches (SL and QP) in Table~\ref{table:combined_rst}. Compared to the baseline (T5-Large PICARD CoSQL), the proposed system achieves significant improvement: we observe 2.1\% and 3.7\% absolute improvement on EM and EX. The table also shows the effect of each component: MT affects the model performance the most on EM and SL leads the most improvement on EX, while QP contributes the least. Note that the overall gains also carry over to T5-3B, with the EM of 58.0\% and EX of 70.0\% representing improvements over the SOTA baseline model. We also submitted our system to CoSQL leaderboard, on the held out test set, PICARD baseline obtained a question match of 54.6\%, while our method achieved 55.8\%. We also improved the interaction match from 23.7\% to 24.8\% with our proposed approach.

\begin{table}[h!]
\vspace{-1mm}
\caption{CoSQL: Baseline, combined results using MT and reranking, and ablations removing one component at a time.}
\vspace{-1mm}
\centering
\label{table:combined_rst}

\begin{tabular}{|l|cc|cc|}
\hline
Method &  \multicolumn{2}{c|}{T5-Large} &  \multicolumn{2}{c|}{T5-3B}   \\
{}   & EM\%   & EX\% & EM\%   & EX\%\\
\hline
Baseline  &   54.4  & 63.7 &   57.1  & 66.6 \\
\hline
Proposed  &   \textit{56.5}  & \textit{67.4} &   \textbf{58.0}  & \textbf{70.0} \\
$\quad -$ MT    &   55.3  & 65.9 &   57.8  & 69.1 \\
$\quad -$ SL    &   56.6  & 65.6 &   58.2  & 69.2 \\
$\quad -$ QP    &   56.7  & 67.2 &   58.1  & 69.5 \\
\hline
\end{tabular}
\end{table}
\vspace{-1mm}

\noindent \textbf{Task difficulty levels}: We now analyze the gains over difficulty levels. Table~\ref{table:gains_diff_levels} shows that the gains from the proposed system (T5-Large model combining MT and reranking) consistently carry over all the pre-defined difficulty levels in CoSQL task. 
\begin{table}[h!]
\caption{T5-large: Performance across difficulty levels on CoSQL.}
\vspace{-2mm}
\centering
\label{table:gains_diff_levels}
\begin{tabular}{|c|c|cc|cc|}
\hline
Diff & count&  \multicolumn{2}{c|}{Baseline} & \multicolumn{2}{c|}{Proposed}\\
{} & {}   & EM\%   &EX\%   & EM\%      &EX\% \\
\hline
Easy   &  417 
        &  74.1   &   77.9 
        &   75.3  &    82.3 \\
Med   &  320  
        &  51.2  &  60.0  
        &  53.8  &  63.8  \\
Hard   &  162 
        &  34.0  &  59.3  
        &  37.7  &  61.1  \\
Extra   &  107  
        &  17.8  &  26.2  
        &  19.6 &  29.9  \\
\hline
Total   &  1006  
        &  54.4  &  63.7 
        &  56.5  &  67.4  \\
\hline
\end{tabular}
\vspace{-1mm}
\end{table}

\noindent \textbf{Effect of additional data on MT}: The proposed MT approach has two parts: a) utilizing additional SParC data compared to baseline; b) attaching task specific prompts to model inputs. For T5-Large, the effect of adding extra data is presented in Table~\ref{table:additional_data_effect} with CD (1.2\% and 0.4\% absolute improvement on EM and EX). Adding task specific prompts give another 1.2\% and 1.3\% improvement on the metrics.

\begin{table}[h!]
\vspace{-1mm}
\caption{T5-large: Additional data effect.}
\vspace{-2mm}
\centering
\label{table:additional_data_effect}
\begin{tabular}{|l|cc|}
\hline
Data  & EM\%   & EX\% \\
\hline

Baseline      &   54.4    &  63.7  \\
W/ extra data        &   55.6   &  64.1  \\
$\quad +$ MT prompt                  & 56.8  & 65.4\\

\hline
\end{tabular}
\vspace{-2mm}
\end{table}

\vspace{-1mm}
\subsection{Turn Level Analysis}
\label{sec:TLA}
\vspace{-2mm}

\begin{figure}[t!]
  \centering
  \includegraphics[width=0.45\textwidth]{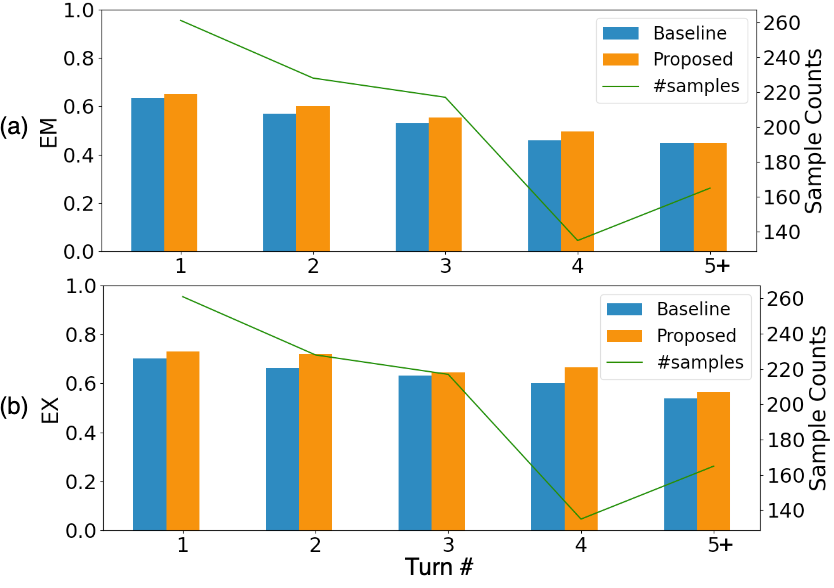}
   \vspace{-2mm}
   \vspace{-2mm}
\caption{T5-large models: Turn level analysis. The y-axes on the left hand side show EM and EX for (a) and (b) respectively, while the right hand side shows sample counts per bin.}
 \vspace{-1mm}
\label{fig:tla}
\end{figure}

Table~\ref{table:Diff_lvl_distri} shows the distribution of difficulty levels across turns. It confirms that later turns have more complex SQL queries. Since the sample counts are extremely small after Turn 5, we merged all subsequent turns into that bin, so that each row in Table~\ref{table:Diff_lvl_distri} has at least 120 examples. These bins are then used in Fig~\ref{fig:tla} to present the performance of the baseline and proposed systems at each turn. The green line shows sample counts of the bins, and numbers could be read from the y-axis on the right side. From the bar chart, we can clearly see the proposed system consistently performs better than the baseline (never worse than the baseline).

\vspace{-2mm}
\begin{table}[h!]
\caption{Distribution of difficulty level at turns.}
\vspace{-2mm}
\centering
\label{table:Diff_lvl_distri}
\begin{tabular}{|c|c|cccc|}
\hline
Turn \# & Total &  \multicolumn{4}{c|}{Difficulty level Distribution (\%)} \\
{} & counts &  Easy   & Medium   & Hard   & Extra Hard \\
\hline
1   &   261 &   49.8    &  27.6 &   14.6    &  8.0      \\
2   &   228 &   39.0    &  36.8 &   15.8    &  8.3      \\
3   &   217 &   38.2    &  35.9 &   16.1    &  9.7      \\
4   &   135 &   37.8    &  29.6 &   17.8    &  14.8     \\
5+  &   165 &   38.8    &  27.9 &   17.6    &  15.8     \\
\hline
\end{tabular}
\vspace{-1mm}
\end{table}

\noindent \textbf{Dialogue Context Dependency}: Some turns are highly dependent on the context provided by previous turns in the dialogue, while others are independent of the context. To understand performance of the proposed system in both cases, we manually annotated each turn in CoSQL DEV set as to whether the context is needed. We then divide the DEV examples into two groups: a) 712 context-independent examples; b) 294 context-dependent examples. Table~\ref{table:context_dep} shows that for T5-Large the proposed system achieves 2.1\% and 3.2\% absolute improvement on EM and EX on the context-independent group, while yielding 2\% and 4.8\% improvement on the context-dependent group. We also present ablations removing one component at a time, with the context-independent group yielding a similar conclusion as in Section~\ref{sec:combined_results}. Meanwhile, the context-dependent group shows that MT contributes the most on both metrics.

\begin{table}[!h]
\caption{T5-large: Context dependency: Performance on CoSQL.}
\vspace{-1.5mm}
\centering
\label{table:context_dep}

\begin{tabular}{|l|cc|cc|}
\hline
Method &   \multicolumn{2}{c|}{Context Independent} &   \multicolumn{2}{c|}{Context Dependent}  \\
{}   & EM\% & EX\% & EM\%   & EX\% \\
\hline
Baseline            &   61.5    &   70.4    &   37.1    &   47.6    \\
\hline
Proposed            &   63.6    &   73.6    &   39.1    &   52.4    \\
$\quad -$ MT        &   62.2    &   72.8    &   38.4    &   49.3    \\
$\quad -$ SL        &   63.9    &   71.6    &   38.8    &   51.0    \\
$\quad -$ QP        &   63.5    &   73.3    &   40.1    &   52.4    \\
\hline
\end{tabular}
\vspace{-4mm}
\end{table}

\vspace{-2mm}
\section{Discussion}
\label{sec:generalization}
\vspace{-1.5mm}

\subsection{Performance on Spider}
\vspace{-1mm}
Table~\ref{table:cross_task_gene} shows the proposed approach also improves over the baseline on the Spider task as well. For example, compared to baseline Spider task-specific model (with T5-Large), we observed 0.6\% and 1.7\% absolute improvement on EM and EX respectively. The T5-Large proposed system even outperforms the SOTA baseline Spider model (a T5-3B model). The table also presents ablations removing one component at a time, showing similar trends as previously in Section~\ref{sec:combined_results}. 

\vspace{-1.5mm}
\begin{table}[h!]
\caption{Cross task generalization: Performance on Spider.}
\vspace{-1.5mm}
\centering
\label{table:cross_task_gene}
\begin{tabular}{|l|cc|}
\hline
Method   & EM\%   & EX\% \\
\hline
Baseline (T5-Large) &   75.1   &  79.4   \\
Baseline (T5-3B)    &   75.6   &  79.3   \\
\hline
Proposed (T5-Large) &   \textbf{75.7}   &  \textbf{81.1}  \\
$\quad -$ MT        &   73.4   &  79.6   \\
$\quad -$ SL        &   75.2   &  79.1   \\
$\quad -$ MT-QP     &   75.0   &  80.9   \\
\hline
\end{tabular}
\vspace{-5.5mm}
\end{table}

\vspace{-2mm}
\subsection{Zero-Shot Generalization (ZSG)}
\vspace{-1mm}
CoSQL and Spider tasks are designed to be cross-domain (without database overlaps over splits), system performances are then reported in zero-shot. However, we wanted to tease apart generalization performance separately due to compositionality and zero-shot domain. To attribute zero-shot ``only'' generalization performance, we ignore DEV examples with parse trees\footnote{To make the parse trees meaningful for this exercise, we prune them so that the leaves for each clause are ignored (values, column and table names).} unseen in MT training data (meaning that all the remaining parse trees were observed in the training data). However, this contains examples with unseen DB schema: we then report ZSG for the proposed and baseline systems in Table~\ref{table:zs}. The performance reported for all systems are now better (because they are slightly easier examples, having removed unseen parse trees). The system generalizes well on both CoSQL and Spider task, with at least 2\% improvement in the metrics at each case. Consistent with previous results, QP contributes the least in this study. A challenge in performing inference on unseen DB schema is the primary/foreign key relationships among tables, such as the example below (note that the turns are concatenated with separator ``$\mid$'').

\begin{figure}[!h]
  \centering
  \vspace{-2mm}
  \includegraphics[width=1\linewidth]{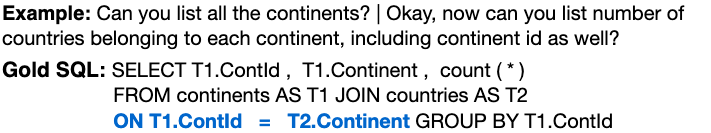}
  \vspace{-8mm}
\end{figure}

\begin{figure}[!h]
  \centering
  \vspace{-4mm}
  \includegraphics[width=1\linewidth]{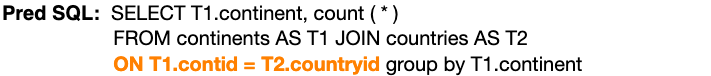}
  \vspace{-5mm}
\end{figure}

\begin{table}[!h]
\vspace{-2.5mm}
\caption{T5-large: Zero-shot ``only'' generalization to unseen DB schema on CoSQL and Spider (using seen parse trees).}
\vspace{-1.5mm}
\centering
\label{table:zs}
\begin{tabular}{|l|cc|cc|}
\hline
Method &  \multicolumn{2}{c|}{CoSQL} &  \multicolumn{2}{c|}{Spider}   \\
{}  & EM\%   & EX\% & EM\%   & EX\%\\
\hline
Baseline  &  59.6    & 68.0 &   81.2    & 85.0\\
\hline
Proposed   & 62.0    & 72.2 &    81.5   &  85.9\\
$\quad -$ MT     &   60.2    & 70.2 &   78.9    & 84.8\\
$\quad -$ SL     & 62.1    & 70.2 &    80.9   &  83.6\\
$\quad -$ QP     & 62.1    & 72.2 &    80.9   &  85.8\\

\hline
\end{tabular}
\vspace{-4mm}
\end{table}

\vspace{1mm}
\subsection{Compositional Generalization (CG)}
\vspace{-1mm}
To analyze CG, we mix the train and DEV sets of the 3 datasets and re-do the split (meaning that the new train and DEV overlap in DB schema, and therefore are not zero-shot). We re-train the baseline and the proposed approach, and then present the systems with held-out examples having unseen parse trees. Table~\ref{table:cg} shows the results: the performance of all systems are poor (even more so on CoSQL, than on Spider), showing that CG is a challenge, especially in conversational multi-turn tasks. The proposed system obtains a ~1\% absolute improvement on EM and EX on both CoSQL and Spider. Based on the ablation studies, QP hurts performance. Below is an example of a SQL with a novel parse tree that the model has to construct now and gets wrong.

\begin{figure}[!h]
  \centering
  \vspace{-2mm}
  \includegraphics[width=1\linewidth]{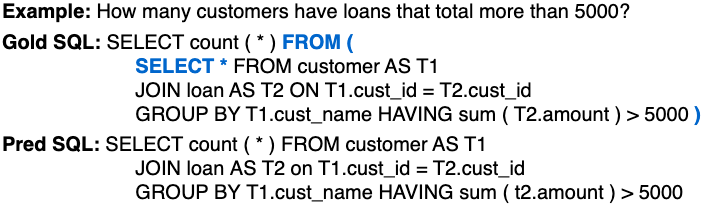}
  \vspace{-5mm}
\end{figure}
\vspace{-3mm}

\begin{table}[!h]
\caption{T5-large: Compositional generalization to unseen parse trees on CoSQL and Spider (using seen DB schema).}
\vspace{-1.5mm}
\centering
\label{table:cg}
\begin{tabular}{|l|cc|cc|}
\hline
Method &   \multicolumn{2}{c|}{CoSQL} &   \multicolumn{2}{c|}{Spider}  \\
{}   & EM\% & EX\% & EM\%   & EX\% \\
\hline
Baseline    &  11.3    &  33.9   &  49.4   &   61.1 \\
\hline
Proposed    &  12.5    &  35.1  &  51.7    &   62.6 \\
$\quad -$ MT      &  8.9    &  35.7   &  48.7    &   60.0 \\
$\quad -$ SL      &  12.5    &  35.1  &  51.7    &   62.6 \\
$\quad -$ QP      &  14.3    &  36.9  &  54.7    &   64.9 \\
\hline
\end{tabular}
\end{table}
\vspace{-2mm}

\vspace{-2mm}
\section{Conclusions}
\vspace{-2mm}
Using task-specific prompts and aggregating coherent task data from Spider, SParC, and CoSQL, we built a T5-family of text-to-SQL models. Generating 10-best lists from this system, and reranking them using query plan model and schema linking algorithm, we achieve significant improvement over the SOTA baseline. Using T5-3B, we obtain absolute improvements of ~$1.0\% $ in EM and $3.4\%$ in EX on the development set over a baseline SOTA system on the competitive CoSQL leaderboard. We achieved consistent improvements on the held out test set of the leaderboard as well. Proposed approach generalizes well to other tasks. Teasing apart generalization performance in terms of zero-shot only and compositionality, while the proposed approach improves over the baseline in both aspects, our study shows that compositionality remains a huge challenge, especially in conversational settings.
\vspace{-2mm}

\bibliographystyle{IEEE}
\bibliography{refs}

\begin{thebibliography}{10}

\bibitem{dahl1994expanding}
Deborah~A Dahl, Madeleine Bates, Michael~K Brown, William~M Fisher, Kate
  Hunicke-Smith, David~S Pallett, Christine Pao, Alexander Rudnicky, and
  Elizabeth Shriberg,
\newblock ``{Expanding the scope of the ATIS task: The ATIS-3 corpus},''
\newblock in {\em Human Language Technology}, 1994.

\bibitem{zelle1996learning}
John~M Zelle and Raymond~J Mooney,
\newblock ``{Learning to parse database queries using inductive logic
  programming},''
\newblock in {\em Proc. of the national conference on artificial intelligence},
  1996.

\bibitem{yu2018spider}
Tao Yu, Rui Zhang, Kai Yang, Michihiro Yasunaga, Dongxu Wang, Zifan Li, James
  Ma, Irene Li, Qingning Yao, Shanelle Roman, Zilin Zhang, and Dragomir Radev,
\newblock ``Spider: A large-scale human-labeled dataset for complex and
  cross-domain semantic parsing and text-to-sql task,''
\newblock in {\em EMNLP}, 2018.

\bibitem{yu2019sparc}
Tao Yu, Rui Zhang, Michihiro Yasunaga, Yi~Chern Tan, Xi~Victoria Lin, Suyi Li,
  Heyang Er, Irene Li, Bo~Pang, Tao Chen, et~al.,
\newblock ``Sparc: Cross-domain semantic parsing in context,''
\newblock in {\em Proc. of ACL}, 2019.

\bibitem{yu2019cosql}
Tao Yu, Rui Zhang, Heyang Er, Suyi Li, Eric Xue, Bo~Pang, Xi~Victoria Lin,
  Yi~Chern Tan, et~al.,
\newblock ``Cosql: A conversational text-to-sql challenge towards cross-domain
  natural language interfaces to databases,''
\newblock in {\em Proc. of EMNLP-IJCNLP}, 2019.

\bibitem{zhong2017seq2sql}
Victor Zhong, Caiming Xiong, and Richard Socher,
\newblock ``{Seq2SQL: Generating structured queries from natural language using
  reinforcement learning},''
\newblock {\em arXiv preprint arXiv:1709.00103}, 2017.

\bibitem{ouyang2022training}
Long Ouyang, Jeffrey Wu, Xu~Jiang, Diogo Almeida, Carroll Wainwright, Pamela
  Mishkin, Chong Zhang, Sandhini Agarwal, Katarina Slama, Alex Gray, et~al.,
\newblock ``Training language models to follow instructions with human
  feedback,''
\newblock in {\em Advances in Neural Information Processing Systems}.

\bibitem{chen2021evaluating}
Mark Chen, Jerry Tworek, Heewoo Jun, Qiming Yuan, Henrique Ponde de~Oliveira
  Pinto, Jared Kaplan, Harri Edwards, Yuri Burda, Nicholas Joseph, Greg
  Brockman, et~al.,
\newblock ``{Evaluating large language models trained on code},''
\newblock {\em arXiv preprint arXiv:2107.03374}, 2021.

\bibitem{li2022competition}
Yujia Li, David Choi, Junyoung Chung, Nate Kushman, Julian Schrittwieser,
  R{\'e}mi Leblond, Tom Eccles, James Keeling, Felix Gimeno, Agustin Dal~Lago,
  et~al.,
\newblock ``Competition-level code generation with alphacode,''
\newblock {\em Science}, vol. 378, no. 6624, 2022.

\bibitem{wang2019rat}
Bailin Wang, Richard Shin, Xiaodong Liu, Oleksandr Polozov, and Matthew
  Richardson,
\newblock ``Rat-sql: Relation-aware schema encoding and linking for text-to-sql
  parsers,''
\newblock in {\em Proc. of ACL}, 2020.

\bibitem{yin2017syntactic}
Pengcheng Yin and Graham Neubig,
\newblock ``A syntactic neural model for general-purpose code generation,''
\newblock in {\em Proc. of ACL (Volume 1: Long Papers)}, 2017.

\bibitem{suhr2020exploring}
Alane Suhr, Ming-Wei Chang, Peter Shaw, and Kenton Lee,
\newblock ``Exploring unexplored generalization challenges for cross-database
  semantic parsing,''
\newblock in {\em Proc. of ACL}, 2020.

\bibitem{lin2020bridging}
Xi~Victoria Lin, Richard Socher, and Caiming Xiong,
\newblock ``Bridging textual and tabular data for cross-domain text-to-sql
  semantic parsing,''
\newblock in {\em Findings of the Association for Computational Linguistics:
  EMNLP 2020}, 2020.

\bibitem{scholak2021picard}
Torsten Scholak, Nathan Schucher, and Dzmitry Bahdanau,
\newblock ``Picard: Parsing incrementally for constrained auto-regressive
  decoding from language models,''
\newblock in {\em Proc. of EMNLP}, 2021.

\bibitem{bordes2012joint}
Antoine Bordes, Xavier Glorot, Jason Weston, and Yoshua Bengio,
\newblock ``Joint learning of words and meaning representations for open-text
  semantic parsing,''
\newblock in {\em Artificial intelligence and statistics}. PMLR, 2012.

\bibitem{nl2sql_paper}
Lu~Zeng, Sree Hari~Krishnan Parthasarathi, and Dilek Hakkani-Tur,
\newblock ``N-best hypotheses reranking for text-to-sql systems,''
\newblock in {\em 2022 IEEE Spoken Language Technology Workshop (SLT)}. IEEE,
  2023.

\bibitem{qin2022survey}
Bowen Qin, Binyuan Hui, Lihan Wang, Min Yang, Jinyang Li, Binhua Li, Ruiying
  Geng, Rongyu Cao, Jian Sun, Luo Si, et~al.,
\newblock ``{A Survey on Text-to-SQL Parsing: Concepts, Methods, and Future
  Directions},''
\newblock {\em arXiv preprint arXiv:2208.13629}, 2022.

\bibitem{hwang2019comprehensive}
Wonseok Hwang, Jinyeong Yim, Seunghyun Park, and Minjoon Seo,
\newblock ``{A comprehensive exploration on WikiSQL with table-aware word
  contextualization},''
\newblock {\em arXiv preprint arXiv:1902.01069}, 2019.

\bibitem{zhang2019editing}
Rui Zhang, Tao Yu, Heyang Er, Sungrok Shim, Eric Xue, Xi~Victoria Lin, Tianze
  Shi, Caiming Xiong, Richard Socher, and Dragomir Radev,
\newblock ``Editing-based sql query generation for cross-domain
  context-dependent questions,''
\newblock in {\em Proc. of EMNLP-IJCNLP}, 2019.

\bibitem{wang2021tracking}
Run-Ze Wang, Zhen-Hua Ling, Jingbo Zhou, and Yu~Hu,
\newblock ``{Tracking interaction states for multi-turn text-to-sql semantic
  parsing},''
\newblock in {\em Proc. of the AAAI Conference on Artificial Intelligence},
  2021, vol.~35.

\bibitem{cai2020igsql}
Yitao Cai and Xiaojun Wan,
\newblock ``Igsql: Database schema interaction graph based neural model for
  context-dependent text-to-sql generation,''
\newblock in {\em Proc. of EMNLP}, 2020.

\bibitem{hui2021dynamic}
Binyuan Hui, Ruiying Geng, Qiyu Ren, Binhua Li, Yongbin Li, Jian Sun, Fei
  Huang, Luo Si, Pengfei Zhu, and Xiaodan Zhu,
\newblock ``{Dynamic hybrid relation exploration network for cross-domain
  context-dependent semantic parsing},''
\newblock in {\em Proc. of the AAAI Conference on Artificial Intelligence},
  2021, vol.~35.

\bibitem{zheng2022hie}
Yanzhao Zheng, Haibin Wang, Baohua Dong, Xingjun Wang, and Changshan Li,
\newblock ``Hie-sql: History information enhanced network for context-dependent
  text-to-sql semantic parsing,''
\newblock in {\em ACL}, 2022.

\bibitem{aghajanyan2021muppet}
Armen Aghajanyan, Anchit Gupta, Akshat Shrivastava, Xilun Chen, Luke
  Zettlemoyer, and Sonal Gupta,
\newblock ``Muppet: Massive multi-task representations with pre-finetuning,''
\newblock in {\em Proc. of EMNLP}, 2021.

\bibitem{xie2022unifiedskg}
Tianbao Xie, Chen~Henry Wu, Peng Shi, Ruiqi Zhong, Torsten Scholak, Michihiro
  Yasunaga, Chien-Sheng Wu, Ming Zhong, Pengcheng Yin, Sida~I. Wang, et~al.,
\newblock ``{U}nified{SKG}: Unifying and multi-tasking structured knowledge
  grounding with text-to-text language models,''
\newblock in {\em Proc. of EMNLP}, 2022.

\bibitem{furrer2020compositional}
Daniel Furrer, Marc van Zee, Nathan Scales, and Nathanael Sch{\"a}rli,
\newblock ``{Compositional generalization in semantic parsing: Pre-training vs.
  specialized architectures},''
\newblock {\em arXiv preprint arXiv:2007.08970}, 2020.

\bibitem{gu2021beyond}
Yu~Gu, Sue Kase, Michelle Vanni, Brian Sadler, Percy Liang, Xifeng Yan, and
  Yu~Su,
\newblock ``{Beyond IID: three levels of generalization for question answering
  on knowledge bases},''
\newblock in {\em Proc. of the Web Conference 2021}, 2021.

\end{thebibliography}

\end{document}